\pdfoutput=1

\documentclass[11pt]{article}

\usepackage{acl}

\usepackage{times}
\usepackage{latexsym}

\usepackage[T1]{fontenc}

\usepackage[utf8]{inputenc}

\usepackage{microtype}

\usepackage{inconsolata}

\usepackage{amsmath}
\usepackage{amssymb}
\usepackage{paralist}
\usepackage{cleveref}

\usepackage{mathtools}
\DeclarePairedDelimiter\abs{\lvert}{\rvert}%
\DeclarePairedDelimiter\norm{\lVert}{\rVert}%
\makeatletter
\let\oldabs\abs
\def\abs{\@ifstar{\oldabs}{\oldabs*}}
\let\oldnorm\norm
\def\norm{\@ifstar{\oldnorm}{\oldnorm*}}
\makeatother

\usepackage[noend,ruled]{algorithm2e}
\crefname{algocf}{algorithm}{algorithms}
\Crefname{algocf}{Algorithm}{Algorithms}

\usepackage{tikz}
\usetikzlibrary {automata,positioning}

\usepackage{xcolor}
\newcommand{\wm}[1]{\textcolor{orange!80}{\textsc{wm}: #1}}
\newcommand{\ye}[1]{\textcolor{purple!40}{\textbf{Yanai: #1}}}
\newcommand{\nascomment}[1]{\textcolor{blue}{\textbf{[#1 --Noah]}}}

\usepackage{listings}
\usepackage{graphicx}
\usepackage{tikz}
\usetikzlibrary{automata, positioning}

\usepackage{booktabs}
\usepackage{adjustbox}
\usepackage{subcaption}
\usepackage{multirow}

\usepackage{enumitem}
\newlist{questions}{enumerate}{2}
\setlist[questions,1]{label=RQ\arabic*.,ref=RQ\arabic*, leftmargin=3em}
\setlist[questions,2]{label=(\alph*),ref=\thequestionsi(\alph*), leftmargin=3em}
\crefformat{questionsi}{#2#1#3}
\Crefformat{questionsi}{#2#1#3}

\usepackage{csquotes}
\renewcommand{\mkbegdispquote}[2]{\itshape}

\newcommand{\dogtitle}{\raisebox{-2pt}{\includegraphics[width=0.22in]{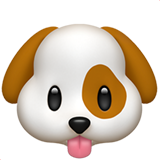}}}

\usepackage{bold-extra}

\usepackage{bbm}

\usepackage{algorithm2e}
\SetKw{Continue}{continue}

\newcommand\open{\textsc{open}}
\newcommand\close{\textsc{close}}

\title{Evaluating $n$-Gram Novelty of Language Models Using \textsc{Rusty-DAWG }\dogtitle}

\definecolor{olmoBlue}{HTML}{265ed4}
\newcommand{\balpha}{{\color{olmoBlue} \boldsymbol{\alpha}}}
\newcommand{\bbeta}{{\color{olmoBlue} \boldsymbol{\beta}}}
\newcommand{\bgamma}{{\color{olmoBlue} \boldsymbol{\gamma}}}

\author{William Merrill$^{\balpha,\bbeta}$ \quad
  Noah A. Smith$^{\bgamma,\bbeta}$ \quad
  Yanai Elazar$^{\bbeta,\bgamma}$\\
  $^\balpha$New York University \quad
  $^\bbeta$Allen Institute for AI \quad
  $^\bgamma$University of Washington \\
  \texttt{willm@nyu.edu}, \texttt{noah@allenai.org}, \texttt{yanaiela@gmail.com}
}

\renewcommand{\wm}[3]{}  
\renewcommand{\ye}[3]{}  
\renewcommand{\nascomment}[3]{}  

\begin{document}

\maketitle

\begin{abstract}%
How novel are texts generated by language models (LMs) relative to their training corpora? In this work, we investigate the extent to which modern LMs generate $n$-grams from their training data, evaluating both (i) the probability LMs assign to complete training $n$-grams and (ii) $n$-novelty, the proportion of $n$-grams generated by an LM that did not appear in the training data (for arbitrarily large $n$). To enable arbitrary-length $n$-gram search over a corpus in constant time w.r.t.~corpus size, we develop \textsc{Rusty-DAWG}, a novel search tool inspired by indexing of genomic data. We compare the novelty of LM-generated text to human-written text and explore factors that affect generation novelty, focusing on the Pythia models. We find that, for $n > 4$, LM-generated text is \emph{less novel} than human-written text, though it is \emph{more novel} for smaller $n$. Larger LMs and more constrained decoding strategies both \emph{decrease novelty}. Finally, we show that LMs complete $n$-grams with lower loss if they are more frequent in the training data. Overall, our results reveal factors influencing the novelty of LM-generated text, and we release \textsc{Rusty-DAWG} to facilitate further pretraining data research.\footnote{\url{https://github.com/viking-sudo-rm/rusty-dawg}}
\end{abstract}

\section{Introduction}


Despite an explosion of new applications of language models (LMs), a core question about LMs as text generators has not been fully answered: \emph{how novel is the text they generate compared to their training corpus?}
This question has both scientific value and practical relevance for LM deployment. From a scientific perspective, language understanding is often theorized as hinging on compositionality, meaning that an infinite range of meanings can be expressed by combining a small set of words or morphemes. If LMs were largely copying sentences or spans they had seen before, this would suggest they cannot compositionally generate new sentences like humans can.
From a societal perspective, the novelty of LM-generated text may also be relevant to legal questions of whether copyrighted materials can be used in LM pretraining data. For instance, a lawsuit between the New York Times and OpenAI (ongoing at the time of writing) hinges on the legal ambiguity of whether including copyrighted material in training data is allowed under fair use \citep{fair-use-blog-2024}. Scientific evaluation of copying behavior in LMs may help guide the resolution of such questions.

In past work, \citet{mccoy2021language} evaluated the novelty of text generated by sampling from small LMs, finding that small $n$-grams in LM-generated text are less novel than in validation text, though larger $n$-grams are more novel.
However, \citet{mccoy2021language}'s LMs were trained on WebText \citep[40 GB;][]{gpt2}, which is 3\% of the size of the Pile \citep[1254 GB;][]{gao2020pile}, which is more representative of recent pretraining datasets.
Thus, it is unclear how their conclusions would transfer to larger-scale, modern LMs.

In this work, we evaluate the $n$-gram generation novelty of LMs of varying sizes trained on \emph{large-scale web data}.
Specifically, we measure the proportion of generated $n$-grams that are novel w.r.t.~the training set across many $n$, which we call $n$-novelty.
Scaling the analysis of $n$-novelty to large corpora is challenging because measuring large-$n$-gram statistics over large corpora is infeasible when implemented naively.
To solve this problem, we develop \textsc{Rusty-DAWG}, a search tool that uses the Compacted Directed Acyclic Word Graph \citep[CDAWG,][]{Crochemore1997,inenaga2005online} data structure for arbitrary-length $n$-gram matching over a corpus in \emph{constant time} w.r.t.~the corpus size and linear w.r.t the query size.
While similar approaches were previously applied to genome data, we develop new algorithms for inference and extending the CDAWG with frequency information, and we are the first, to our knowledge, to scale up CDAWGs to LM training data.
We use \textsc{Rusty-DAWG} to address the following research questions, focusing on the Pythia suite of LMs \citep{biderman2023pythia} trained on the Pile:
\begin{questions}
    \item How novel is typical text generated by LMs compared to new human-written text? \label{rq:absolute-novelty}
    \item How do model size, decoding strategies, and prompting with training data influence the novelty of model-generated text? \label{rq:relative-novelty}
    \item Across $n$-gram sizes, how does the occurrence and frequency of $n$-grams in the training set impact their completion loss? \label{rq:completion-loss}
\end{questions}
\noindent Our contributions and findings are as follows:
\begin{enumerate}
    \setcounter{enumi}{-1}
    \item We introduce \textbf{\textsc{Rusty-DAWG}},
    an efficient data structure based on CDAWG automata that enables unbounded-length $n$-gram searches in massive pretraining datasets.
    \item We find {\bf large $n$-grams ($n > 4$) are less novel} in LM-generated text compared to human-written text, though small $n$-grams ($n \leq 4$) are more novel (\Cref{rq:absolute-novelty}, \Cref{sec:absolute-novelty}).
    \item We show that \textbf{novelty decreases with larger LMs and constrained decoding} (\Cref{rq:relative-novelty}, \Cref{sec:relative-novelty}). To an extent, prompting with training data also decreases novelty.
    \item We show \textbf{LMs complete
    frequent training $n$-grams with lower loss} (\Cref{rq:completion-loss}, \Cref{sec:completion-loss}).
\end{enumerate}


\section{Operationalizing Novelty with $n$-Grams} \label{sec:novelty-def}

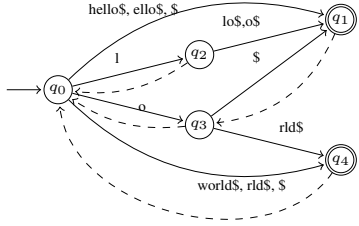
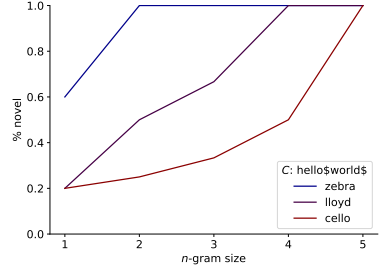
\begin{figure*}
    \centering
    \begin{subfigure}[t]{0.4\linewidth}
        \centering
        \includegraphics[width=1.\linewidth]{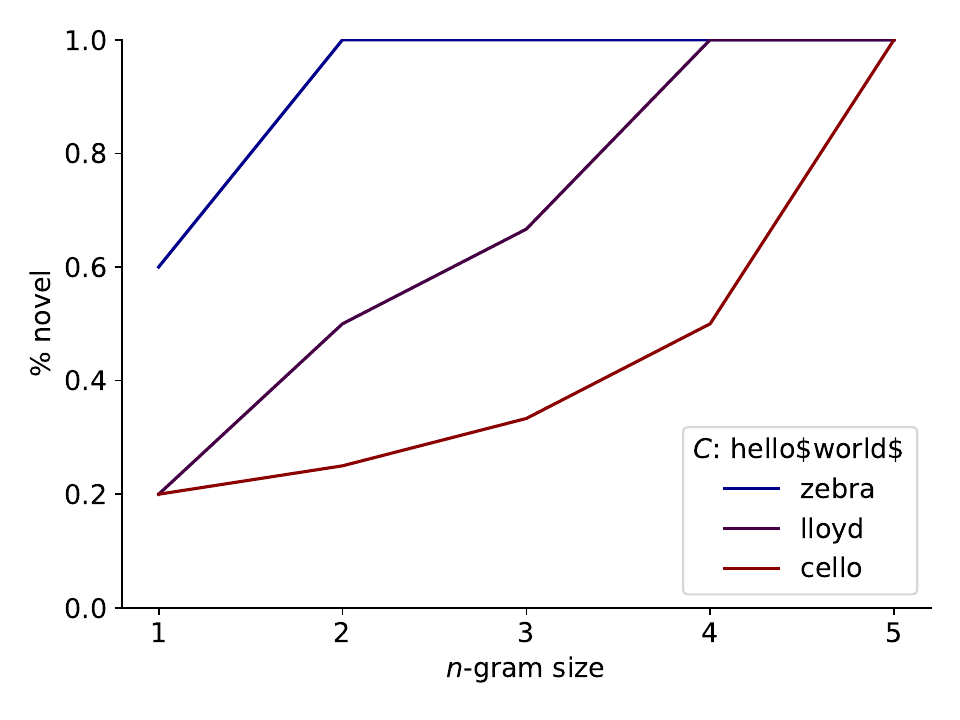}
        \caption{Novelty curves computed from the CDAWG in \Cref{fig:example-cdawg}, labeled by their corresponding queries.}
        \label{fig:example-novelty}
    \end{subfigure}
    \qquad
    \begin{subfigure}[t]{0.4\linewidth}
        \centering
        \resizebox{1.\columnwidth}{!}{
        \begin{tikzpicture}[shorten >=1pt, node distance=2cm, on grid, auto, initial text=,font=\tiny]

\tikzset{state/.style={
    circle,
    draw,
    minimum size=0.4cm, 
    inner sep=1pt 
}}

\node[state, initial] (q0) at (0, 0) {$q_0$};
\node[state, accepting] (q1) at (4, 1) {$q_1$};
\node[state] (q2) at (2, .5) {$q_2$};
\node[state, accepting] (q4) at (4, -1) {$q_4$};
\node[state] (q3) at (2, -.5) {$q_3$};

\path[->]
(q0) edge[bend left] node[left] {hello\$, ello\$, \$} (q1)
(q0) edge node {l} (q2)
(q0) edge  node[right] {o} (q3)
(q2) edge  node {lo\$,o\$} (q1)
(q3) edge  node {\$} (q1)
(q3) edge  node {rld\$} (q4)
(q0) edge[bend right] node[below right] {world\$, rld\$, d\$, \$} (q4)
(q1) edge[bend left, dashed] node {} (q3)
(q2) edge[bend left, dashed, in=160, out=20] node {} (q0)
(q3) edge[bend left, dashed, in=160, out=20] node {} (q0)
(q4) edge[bend left, dashed, in=110, out=70] node {} (q0)
;
\end{tikzpicture}
        }
        \caption{CDAWG for $C=\texttt{hello\$world\$}$, where $\texttt{\$}$ is a document separator. Dashed arrows are failure arcs.}
        \label{fig:example-cdawg}
    \end{subfigure}
   \caption{Illustration of CDAWG and resulting novelty curves with character-level tokenization for simplicity.} \label{fig:example}
\end{figure*}

There are different ways to measure LM generation novelty:
one could assess the verbatim overlap between the text and training data \citep{mccoy2021language} or attempt to capture semantic and syntactic novelty \cite{shaib2024standardizing,syntactic-templates}.
We target verbatim novelty, which is understudied at large model scale and conceivably provides an upper bound on more semantic notions of novelty (any instance of verbatim repetition likely also constitutes semantic repetition).
We formalize verbatim novelty via two $n$-gram-based metrics: {\bf $n$-novelty} and \textbf{non-novel suffix length (\textsc{nnsl})}.
\paragraph{$n$-Novelty.} Novelty can be evaluated at different levels of granularity. For example, while all individual tokens in a generated text will likely have occurred, it would be notable if a 100-gram from the pretraining data was generated verbatim.
To capture novelty across different $n$-gram lengths, we follow \citet{mccoy2021language} in plotting the $n$-novelty curve, i.e., the novelty of generated $n$-grams (where $n$ varies) w.r.t. some fixed corpus $C$.
Formally, for any text query $Q$ (e.g., a model-generated document)
we define the $n$-novelty rate of $Q$ as the proportion of $n$-grams in $Q$ that did not occur in $C$.
We visualize the $n$-novelty curve as a function of $n$ in \Cref{fig:example-novelty}.  Intuitively, 1-novelty should be close to zero (due to the way tokenizers work), and the curve will monotonically increase with $n$ (since substrings of a non-novel $n$-gram are non-novel).
\paragraph{\textsc{nnsl}.} We propose a new measure of aggregate novelty across different values of $n$. We define \textsc{nnsl} at token $i$ of $Q$ as the length of the longest suffix of $Q[:i]$ that appeared in corpus $C$. We then aggregate with mean or max.

\paragraph{Example.}
Let $C = \texttt{hello\$world\$}$ be a character-tokenized corpus, where $\texttt{\$}$ is a document boundary. Query $Q = \texttt{lloyd}$ has 1-gram novelty 1/5 (\texttt{y} is novel), 2-gram novelty 2/4 (\texttt{oy} and \texttt{yd} are novel), 3-gram novelty 2/3 (only \texttt{llo} is non-novel), and 4-gram novelty 2/2. The \textsc{nnsl} at each position is $\langle 1, 2, 3, 0, 1\rangle$, with mean 1.4 and max 3.
We intuitively demonstrate this example in \Cref{fig:example-novelty}.

\section{Measuring Novelty with CDAWGs} \label{sec:cdawgs}

Naively computing our novelty metrics is prohibitively expensive over a large pretraining corpus like the Pile (334B tokens).
To make the searches fast, we use a Compacted Directed Acyclic Word Graph \citep[CDAWG;][]{Crochemore1997,inenaga2005online}, a data structure
which returns the \textsc{nnsl} at each position in $Q$ against $C$ in \emph{constant time} (w.r.t.~the size of $C$), and \emph{linear time} (w.r.t. the size of $Q$), from which $n$-novelty can be computed. We describe how to compute \textsc{nssl} using CDAWG in \Cref{sec:computing-novelty}.
This constant-time querying is crucial for our application of searching the Pile.
We first discuss querying CDAWGs (\Cref{sec:querying}), then their memory costs (\Cref{sec:memory}), their construction (\Cref{sec:building}), and our open-source implementation (\Cref{sec:rusty}).

\subsection{Querying CDAWGs} \label{sec:querying}

A CDAWG is a finite-state machine built for a corpus $C$ that acts as a rich index for $C$ (see \Cref{fig:example-cdawg}).
In \Cref{sec:cdawg-query}, we propose an algorithm where a CDAWG for $C$ can be used to efficiently answer \textbf{\textsc{nnsl} queries} on $C$:
\begin{compactitem}
    \item \textsc{Input:} A string (e.g., $Q = \texttt{lloyd}$).
    \item \textsc{Output:} \textsc{nnsl} at each position in $Q$ (e.g., $L(Q) = \langle 1, 2, 3, 0, 1 \rangle$) as well as the training frequencies of the largest suffixes matched at each position (e.g., $N(Q) = \langle 3, 1, 1, 0, 1 \rangle$).
\end{compactitem}
The algorithm follows a single arc for each token in $Q$, maintaining a state that encodes the largest currently matched suffix.
This takes time $O(\abs{Q})$ with \emph{no dependence} on $\abs{C}$, which makes CDAWGs useful faster than suffix arrays \citep{carlini2023quantifying,liu2024infinigram} for searching large corpora.

For illustration (\Cref{fig:example}), we use character-level tokenization, but this process can be applied with any tokenization.
The $n$-novelty curve, as well as all other data presented in this paper, can be computed from non-novel suffix queries.

\subsection{Memory Overhead} \label{sec:memory}

A practical concern for an indexing data structure is its memory overhead: how many bytes does it use on a corpus of size $\abs{C}$?
The CDAWG refines the earlier Directed Acyclic Word Graph \citep[DAWG;][]{blumer1984building} to reduce memory overhead.
A DAWG contains at most $2 \abs{C}$ states and $3 \abs{C}$ arcs \citep{blumer1984building}, which, while linear, becomes impractical for large datasets.
In contrast, a CDAWG achieves $0.18 \abs{C}$ states and $0.97 \abs{C}$ arcs  on the Pile. As a result, we find the CDAWG takes $\sim$50\% as much memory to store as the vanilla DAWG in practice.\footnote{A CDAWG arc is larger than a DAWG arc. Hence, the CDAWG memory overhead is reduced by 50\% (and not more) despite a larger reduction in the number of states and arcs.}
Still, the CDAWG takes $29 \abs{C}$ bytes vs.~$7\abs{C}$ for a suffix array, illustrating a time/space tradeoff between the two approaches.

Another factor that affects memory overhead is the choice of graph representation. We implemented the edge list for a node with an AVL tree to make transitions very fast, but at the cost of some memory overhead.
Further details about the graph representation, memory overhead, and potential improvements can be found in \Cref{sec:cdawg-memory}.


\subsection{Building CDAWGs} \label{sec:building} 

The naive way to build a CDAWG would involve enumerating all span in a corpus in quadratic time, which is infeasible for large corpora.
Luckily, more refined algorithms for building DAWGs and CDAWGs were developed that process each token in the corpus left to right, taking linear time overall \citep{blumer1984building,Crochemore1997,inenaga2005online}.
We implement \citet{inenaga2005online}'s linear-time algorithm.

In \Cref{sec:cdawg-counts}, we propose a graph-traversal algorithm that can be used to add $n$-gram frequency information to the CDAWG after it has been created, stored at the states.
Since edges dominate the memory overhead of the CDAWG, this only minimally increases the space overhead but allows us to return the count associated with matched $n$-grams, which we expect to be useful in many applications.

\subsection{\textsc{Rusty-DAWG} Library}
\label{sec:rusty}
While there are some pre-existing open-source libraries for DAWGs,\footnote{\url{https://github.com/elake/SuffixAutomaton}}
we did not find a scalable open-source implementation of CDAWGs.
To facilitate our research and other applications of CDAWGs to large text corpora, we implemented \textsc{Rusty-DAWG}, a modern Rust library for building and using DAWGs and CDAWGs, which we open-sourced at: \url{https://github.com/viking-sudo-rm/rusty-dawg}.
\textsc{Rusty-DAWG} provides a memory-efficient implementation of CDAWGs, their construction algorithm, and our algorithms for populating them with frequency information and answering \textsc{nnsl} queries.
It also provides useful features like the ability to store the CDAWG in RAM or on disk (and switch between these modes).
\textsc{Rusty-DAWG} enabled us to build, to our knowledge, the largest CDAWG every created.
See \Cref{sec:cdawg-library} for more details.

\section{Experimental Setup} \label{sec:experimental-setup}

\subsection{Building a CDAWG on the Pile}

We focus our study on the copying behavior of the eight Pythia models \citep{biderman2023pythia} trained on the Pile \citep{gao2020pile}.
The Pile contains many kinds of text, including web text, books, code, and email communication.
We build our \textsc{Rusty-DAWG} on the non-deduplicated version using the GPT-NeoX \citep{black-etal-2022-gpt} tokenization used by Pythia, under which it contains 334B tokens.

To parallelize building \textsc{Rusty-DAWG},
we divide the Pile documents into 30 shards and build a CDAWG on each 11B-token shard on a different cloud machine. Each of the 30 created CDAWGs has 2B states and 11B arcs, taking 327 GB total memory. We store this in RAM during building.
At inference time, we keep the CDAWG shards on disk and execute \textsc{nnsl} queries on each of the 30 CDAWGs in parallel. We aggregate \textsc{nnsl} (by taking the max) and counts returned (by summing at maximum suffix lengths) to exactly simulate the output of a single CDAWG.

\subsection{Generating Text from LMs}

We evaluate the generation novelty of the Pythia models \citep{biderman2023pythia}, which were trained on the Pile \citep{gao2020pile} at different sizes up to 12B parameters.
We consider two setups, (1) generating unmprompted texts, and (2) generating prompted texts, for which we sample 500 documents from the Pile validation set (trimmed to 1,000 tokens).
In each setup we generate 500 documents of 1,000 tokens from each LM.
We vary the model size (from 70M to 12B, 8 models in total) and decoding strategy, sweeping different parameters for top-$p$ \citep[nucleus sampling;][]{Holtzman2020The}, top-$k$ \citep{fan-etal-2018-hierarchical}, temperature, and greedy beam search.
Unless otherwise indicated, we use Pythia-12B and standard sampling with an unconditioned prompt as defaults.
We pass each generated text through the CDAWG to compute the \textsc{nnsl} at each position (cf. \Cref{sec:cdawgs}), from which we compute the $n$-novelty curves.

\subsection{Novelty Baselines} \label{sec:baselines}

For small $n$, some $n$-grams will likely be repeated between a document and a large corpus by random chance. For large $n$ this probability will decrease rapidly.
Thus, to evaluate the novelty of LM generations, it is necessary to establish a baseline $n$-novelty curve.
We consider two such baselines:

\textbf{Validation Text.} Following \citet{mccoy2021language}, we use the novelty of text in the Pile's validation set as a baseline. If $n$-grams of a certain size are less novel in generated text compared to validation text, the LM is generating pretraining $n$-grams more commonly than expected for new documents from the pretraining distribution. This suggests the LM is copying from its pretraining corpus.

\textbf{Text After Pile Date Cutoff.} The novelty of validation text may be artificially
low if the training distribution contains duplicated documents \citep{lee-etal-2022-deduplicating}.
To account for this, we filter text from Dolma \citep{dolma} that was written after the Pile collection cutoff.
Specifically, the two domains we use are Reddit and scientific texts (Pes2o; \citealp{peS2o}), both of which are in-distribution for the Pile.
Thus, we expect this baseline to represent natural overlap for human-written text without contamination.
We report the $n$-novelty curve fit on both domains from Dolma together, though qualitatively we observe that the curve looks similar within each domain.

For both the Pile and Dolma, we sample 500 validation documents and truncate each document to the first 1,000 tokens.

\section{Novelty of LM-Generated Text}
\label{sec:novelty}


\subsection{Novelty vs. Human-Written Text} \label{sec:absolute-novelty}

We now compare the novelty of typical LM-generated text against human text (\Cref{rq:absolute-novelty}).
As such, we
report novelty metrics for two human-written text baselines: validation text and Dolma documents written after the Pile cutoff.

\begin{figure}[t]
    \centering
    \includegraphics[width=1.\columnwidth]{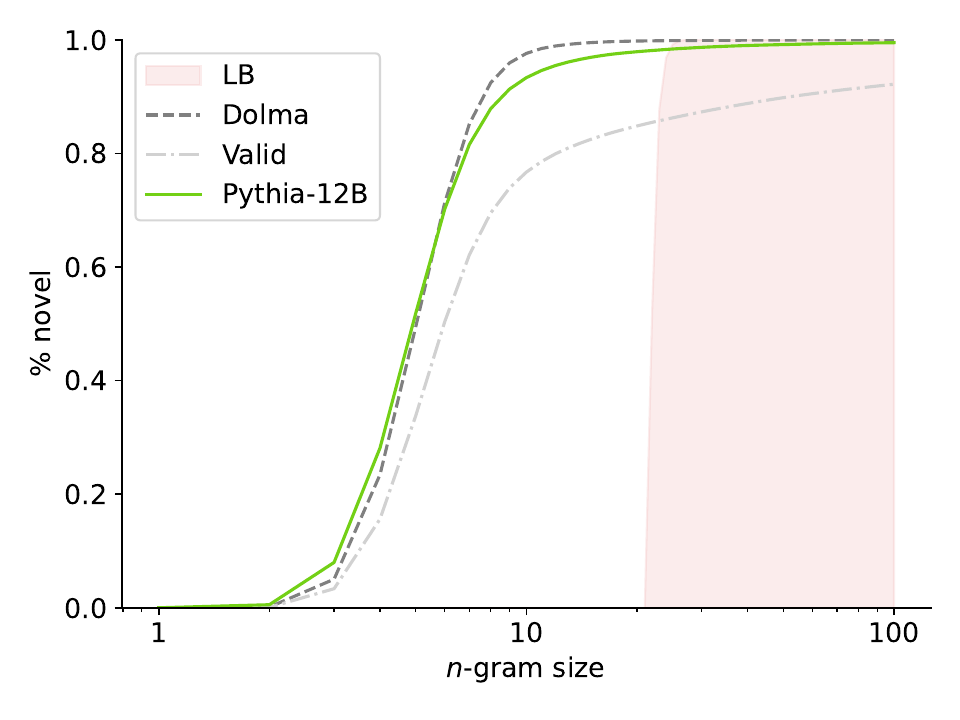}
    \caption{$n$-novelty curve for Pythia-12B with naive sampling. Compared to Dolma, LM-generated text is more novel for $n > 4$ and slightly less novel for $n\leq4$. The gap between the dark gray Dolma curve and the green Pythia-12B curve quantifies the novelty difference. LM-generated text is more novel than the Pile validation set across $n$-gram sizes due to contamination.}
    \label{fig:main}
\end{figure}

\begin{figure}[t]
    \centering
    \begin{subfigure}[t]{0.9\linewidth}
         \centering
         \includegraphics[width=1.\columnwidth]{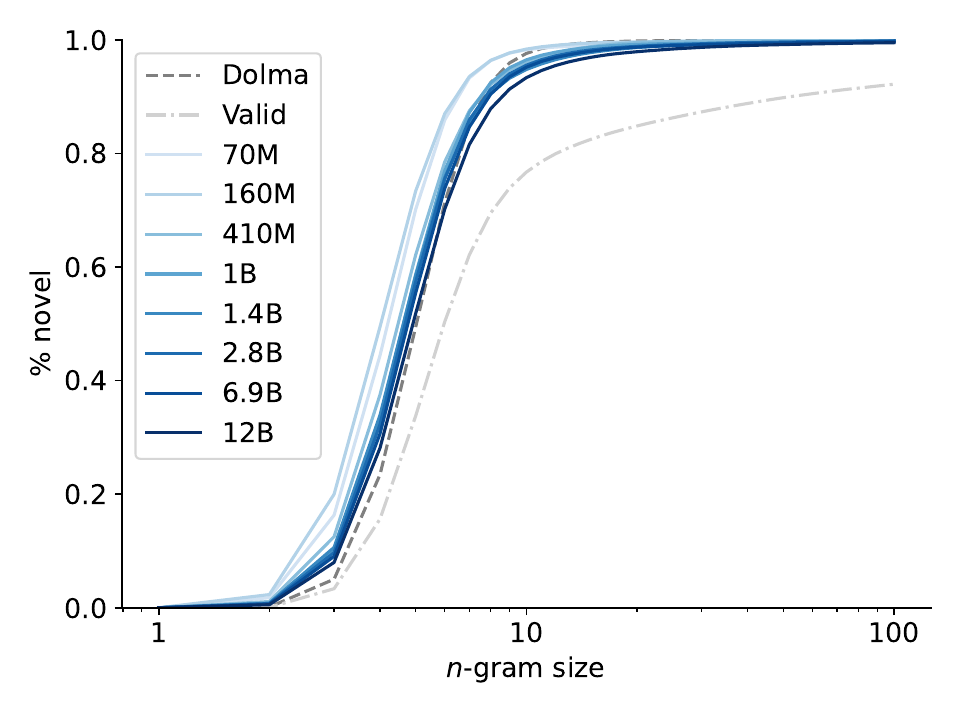}
         \caption{$n$-novelty curves across model sizes.}
         \label{fig:by-model-novelty}
     \end{subfigure}
     \begin{subfigure}[t]{0.9\linewidth}
         \centering
         \includegraphics[width=1.\columnwidth]{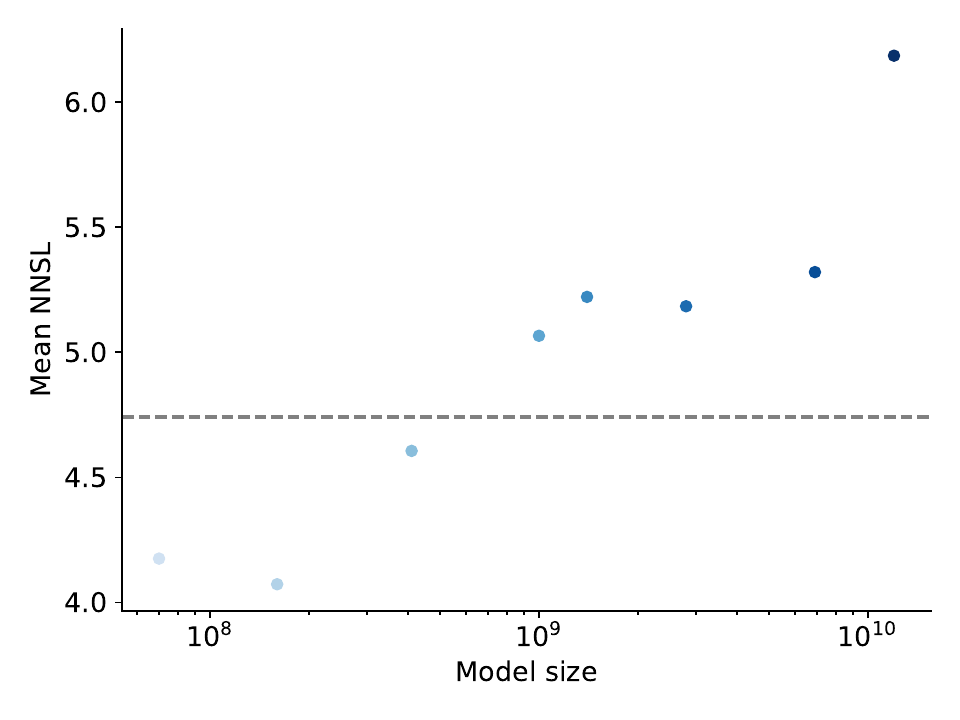}
         \caption{Mean \textsc{nnsl} across model sizes.}
         \label{fig:by-model-nnsl}
     \end{subfigure}
    \caption{Both $n$-novelty and mean \textsc{nnsl} suggest larger LMs generate less novel text than smaller LMs.
    }
    \label{fig:by-model}
\end{figure}

\paragraph{Validation Baseline.}
\Cref{fig:main} shows that the validation $n$-novelty curve is very low across $n$.
2.4\% of the 1,000-token validation documents are \emph{exactly matched} somewhere in the Pile training set.
13.6\% share a 100-gram with the training data, and 25.0\% share a 50-gram.
This suggests contamination, since we expect natural large-$n$-gram overlap should be vanishingly unlikely.

To formally test this, we derive a lower bound on $n$-novelty assuming most next tokens are nondeterministic (\Cref{sec:theoretical-lb}).\footnote{We make the possibly strong assumption that 90\% of tokens per $n$-gram have entropy $\ell \geq 1.8$ bits/token based on the best achieved Pile losses \citep{du2022glm}. See \Cref{sec:theoretical-lb}.}
Under this assumption, the $n$-novelty curve for non-contaminated data should not enter the red region in \Cref{fig:main}, i.e., almost all $23{+}$-grams should be novel.
The validation curve (but not Dolma) enters this region, suggesting many contaminated $n$-grams in the validation text.
Thus, we turn to Dolma as a better representation of uncontaminated human-written text.

\paragraph{Dolma Baseline.}
\Cref{fig:main} shows that $n$-grams of size $n > 4$ are less novel in generated text compared to Dolma text, whereas $n$-grams of size $n \leq 4$ (median length) are slightly more novel.
For instance, 8\% of Pythia bigrams are novel (vs.~5\% for Dolma), while 93\% of Pythia 10-grams and 99\% of 100-grams are novel (vs.~98\% and 100\%).
This disagrees with \citet{mccoy2021language}'s findings for small LMs trained on 40 GB of text, where LMs were less novel for small $n$-grams (2-3\% bigram novelty for LMs vs. 6\% baseline novelty). One explanation for the difference may be the model and data scale, motivating us to more closely analyze the impact of model size on novelty in \Cref{sec:relative-novelty}.

\paragraph{Examples of Copied $n$-Grams.}

We find that many non-novel $n$-grams generated by Pythia-12B are pieces of licenses and boilerplate code. For example, Pythia-12B generates a 64-gram with 45K occurrences in the Pile that starts:
\begin{displayquote}
//\\
//  Licensed under the Apache License, Version 2.0 (the "License");\\
//  you may not use this file except in compliance with the License$\ldots$
\end{displayquote}

\noindent Another generated 64-gram (with 213K occurrences in the Pile) imports Linux libraries:

\begin{displayquote}

$\ldots$\#include <sound/core.h>\\
\#include <sound/pcm.h>\\
\#include <sound/soc.h>$\ldots$


\end{displayquote}

\subsection{Impact of Model Size and Decoding} \label{sec:relative-novelty}

Having explored the novelty of LM-generated text compared to human-written text, we assess the factors that influence the generation novelty of LMs (\Cref{rq:relative-novelty}).
We compare $n$-novelty curves for varying model sizes, decoding strategies, and different prompt lengths from the training data.
The default values of each variable (when sweeping other variables) are 12B, standard sampling, and a prompt length of 0.

\paragraph{Larger LMs are Less Novel.}
\Cref{fig:by-model-novelty} shows that, across $n$, $n$-grams are less novel for larger LMs than for smaller LMs. Similarly, \Cref{fig:by-model-nnsl} shows that the mean \textsc{nnsl} increases linearly with log model size.
Both metrics suggest that larger LMs are \emph{less novel} than smaller LMs across all $n$-gram sizes. This may indicate that larger LMs have more capacity to memorize $n$-grams from training.

\begin{table}[t!]
\centering
\begin{adjustbox}{max width=.8\columnwidth}
\begin{tabular}{llrr}
\toprule

Setup & Param & Mean & Max \\
\midrule
\multirow{2}{*}{Baseline}
& Validation & 29.94 & 1,000 \\
& Dolma & 4.74 & 66 \\

\midrule

\multirow{8}{*}{Size}
& 70M & 4.18 & 187 \\
& 160M & 4.07 & 207 \\
& 410M & 4.61 & 191 \\
& 1B & 5.07 & 270 \\
& 1.4B & 5.22 & 225 \\
& 2.8B & 5.18 & 322 \\
& 6.9B & 5.32 & 198 \\
& 12B & 6.19 & 376 \\

\midrule

\multirow{3}{*}{Prompt}
& 1 & 5.83 & 624 \\
& 10 & 6.21 & 393 \\
& 100 & 7.56 & 976 \\
\bottomrule
\end{tabular}
\end{adjustbox}
\caption{\textsc{nnsl} results for human-written text baselines and different model sizes and prompt lengths.}
\label{tab:nnsl-main}
\end{table}
\begin{table}[t]
\centering
\begin{adjustbox}{max width=.8\columnwidth}
\begin{tabular}{llrr}
\toprule
Decoding & Param & Mean & Max \\
\midrule

\multirow{2}{*}{Baseline}
& Validation & 29.94 & 1,000 \\
& Dolma & 4.74 & 66 \\
\midrule

\multirow{3}{*}{Top-$p$}
& 0.85 & 15.02 & 992 \\
& 0.9 & 8.85 & 1000 \\
& 0.95 & 9.69 & 902 \\

\midrule

\multirow{3}{*}{Top-$k$}
& 20 & 11.34 & 507 \\
& 80 & 9.24 & 580 \\
& 160 & 8.17 & 386 \\

\midrule

\multirow{6}{*}{Temperature}
& 0.5 & 14.22 & 983 \\
& 0.85 & 10.18 & 969 \\
& 0.9 & 11.05 & 1,000 \\
& 0.95 & 6.55 & 418 \\
& 1.05 & 5.08 & 313 \\
& 1.1 & 4.34 & 375 \\

\midrule

\multirow{3}{*}{Beam}
& 8 & 192.03 & 408 \\
& 4 & 9.17 & 18 \\
& 1 & 8.40 & 19 \\
\bottomrule
\end{tabular}
\end{adjustbox}
\caption{\textsc{nnsl} results for Pythia-12B with different decoding strategies. Across strategies, more constrained decoding leads to less novel text.}
\label{tab:nnsl-decoding}
\end{table}

\paragraph{Decoding Constraints Decrease Novelty.}
Prior work with small LMs and corpora suggests decoding choices could influence generation novelty \citep{mccoy2021language}. In particular, we expect more constrained decoding to decrease novelty \citep{liu2024infinigram}.
To evaluate this, we generate text with top-$p$, top-$k$,
temperature (including greedy), and greedy beam search decoding setups, varying the parameter that constraints generation in each case. We hypothesize that the parameter choices that more constrained will result in lower generation novelty.

\begin{figure*}[!t]
    \centering
    \includegraphics[width=0.95\columnwidth]{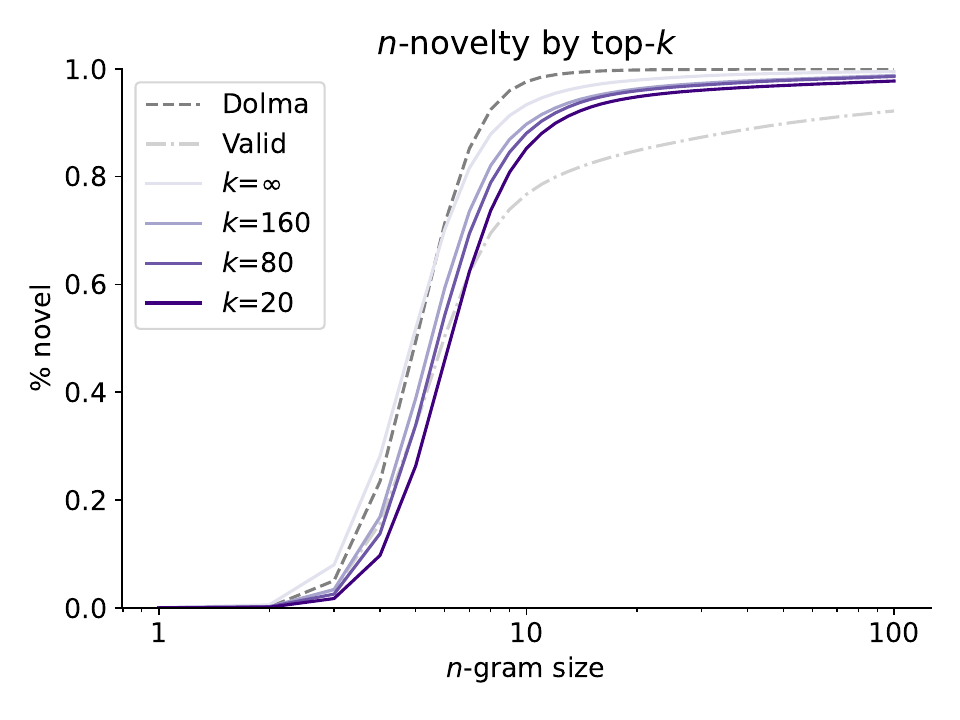}
    \includegraphics[width=0.95\columnwidth]{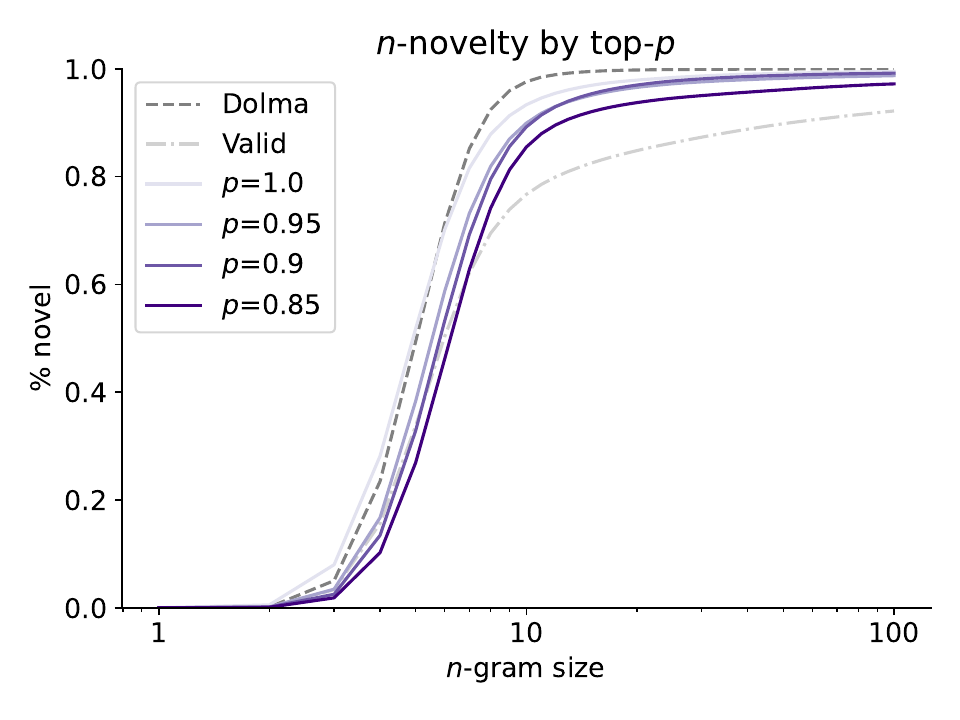}
    \includegraphics[width=0.95\columnwidth]{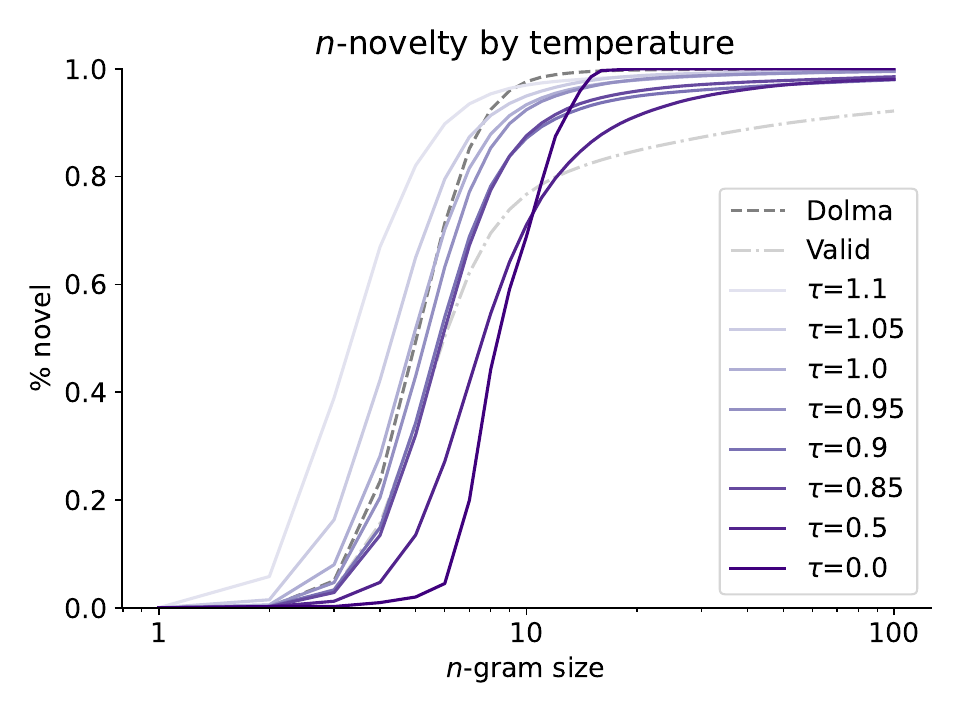}
    \includegraphics[width=0.95\columnwidth]{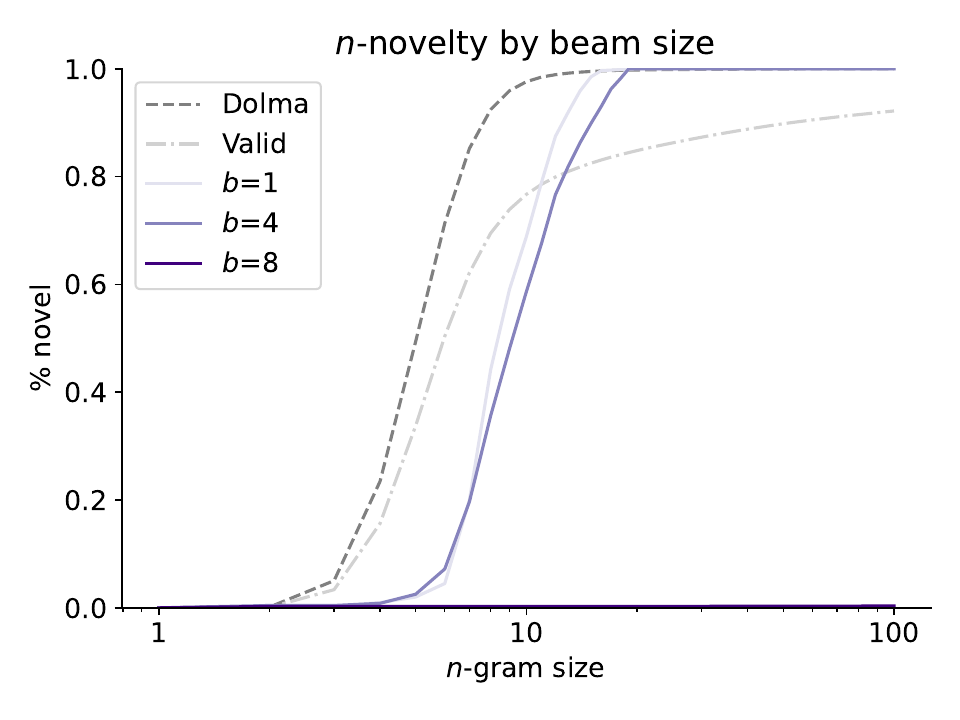}
    \caption{Impact of decoding choices (top-$k$, top-$p$, temperature $\tau$, and beam size with $\tau=0$) on $n$-novelty. Less stochastic (darker) decoding choices \emph{decrease novelty}; temperature and beam size have the strongest effect.}
    \label{fig:by-decoding}
\end{figure*}

Indeed, \Cref{fig:by-decoding} shows that constrained decoding reduces $n$-novelty.
The constrained decoding curves are consistently below the Dolma baseline, and, for small $n$, even below the validation baseline.
The least $n$-novel approaches are low-temperature decoding and beam search.
For 10-grams, temperature 0.5 reaches 71\% novelty and temperature 0 reaches 69\% novelty,
while for 100-grams, temperature 0.5 reaches 98\% and temperature 0 reaches 100\%.
Increasing beam size decreases novelty, with beam size 8 remaining near 0\% novelty even up to 100-grams. With beam size 8, the LM deterministically generates a single document containing a 408-gram from training (see \Cref{app:beam-novelty} for beam size results with nondeterministic conditioned generation).
As we show in \Cref{tab:nnsl-decoding}, temperature $\leq 0.9 $ and top-$p \leq 0.95$ dramatically increase both the mean \textsc{nnsl} and the max.
Both novelty metrics suggest that constrained decoding reduces generation novelty.

\paragraph{Long Training Prompts Slightly Decrease Novelty.} To evaluate the impact of prompting with training data, we prompt the model with $p$ tokens from the beginning of a training document before generating 1,000 additional tokens. We then evaluate the $n$-novelty curve for these 1,000 tokens. Qualitative inspection reveal that the novelty curves look almost identical, independently of the prompt length $p$. However, the \textsc{nnsl} statistics (\Cref{tab:nnsl-main}) tell a more subtle story: the median \textsc{nnsl} remains unchanged, whereas the mean increases from 6.19 to 7.56 with 100 prompt tokens. This suggests that, while most $n$-grams do not become more novel when a prompt is given, the longest non-novel $n$-grams are longer when a longer prompt is given.

\section{Impact of $n$-Gram Training Frequency} \label{sec:completion-loss}

Finally, regarding \Cref{rq:completion-loss}, we test whether, at inference time, LMs assign higher probability to $n$-grams from training, and how this interacts with training frequency.
We define the \textbf{mean completion loss} of $x_1 \cdots x_n$ as the average probability the LM assigns to $x_n$ when it occurs in validation text:
\begin{equation*}
    \hat \ell(x) = \frac{1}{\abs{\mathcal V_x}} \sum_{i \in \mathcal V_x} p_\mathrm{LM}(v_i \mid v_{1:i - 1}) ,
\end{equation*}
\noindent where $\mathcal V_x = \{ i : v_{i+1-n:i} = x \}$.
This captures the LM's sensitivity to training $n$-grams in a way that is independent of the specific sampling choices made when decoding from the LM.
It also captures use cases of LMs where the LM is used to assign probabilities to strings rather than as a text generator, such as in multiple-choice question answering like MMLU \citep{hendryckstest2021} or evaluation of noun-verb agreement \citep{marvin-linzen-2018-targeted}.

\paragraph{Method.} We sample 5,000 documents of 1,000 tokens each, from the Pile validation set. We compute the per-token loss using Pythia-12B and use the CDAWG to find the non-novel suffixes at each position.
For each $n$, we find tokens in the validation data that fall into two categories:
\begin{compactitem}
    \item \textbf{In Train:} The $n$-gram ending at the token occurred in the training data.
    \item \textbf{Not in Train:} The $n$-gram ending at the token did not occur in the training data, but the $(n-1)$-gram ending at the previous token did.
\end{compactitem}
We then compute the mean completion loss across all tokens in each condition with the same value of $n$, and plot this mean loss as a function of $n$. This quantity measures the surprisal of the LM when completing $n$-grams, with the two conditions differentiating whether the correct $n$-gram completion appeared in the training data. For the $n$-grams in the training data, we also investigate how their frequency affects completion loss.

\begin{figure*}[t!]
    \centering
    \begin{subfigure}{0.4\linewidth}
        \centering
        \includegraphics[width=1.\linewidth]{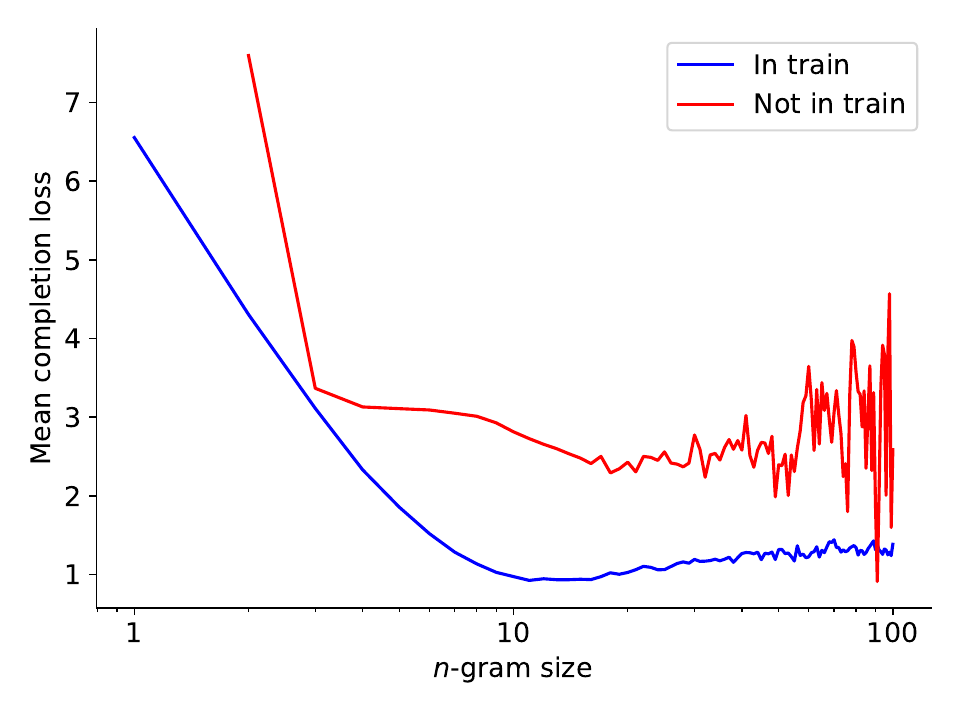}
        \caption{Completion loss of Pythia-12B on $n$-grams in validation text based on whether the $n$-grams occurred in training. Across $n$-gram sizes, Pythia-12B assigns lower loss to $n$-grams seen during training.}
        \label{fig:cl-occurrence}
    \end{subfigure}
    \qquad
    \begin{subfigure}{0.4\linewidth}
        \centering
        \includegraphics[width=1.\linewidth]{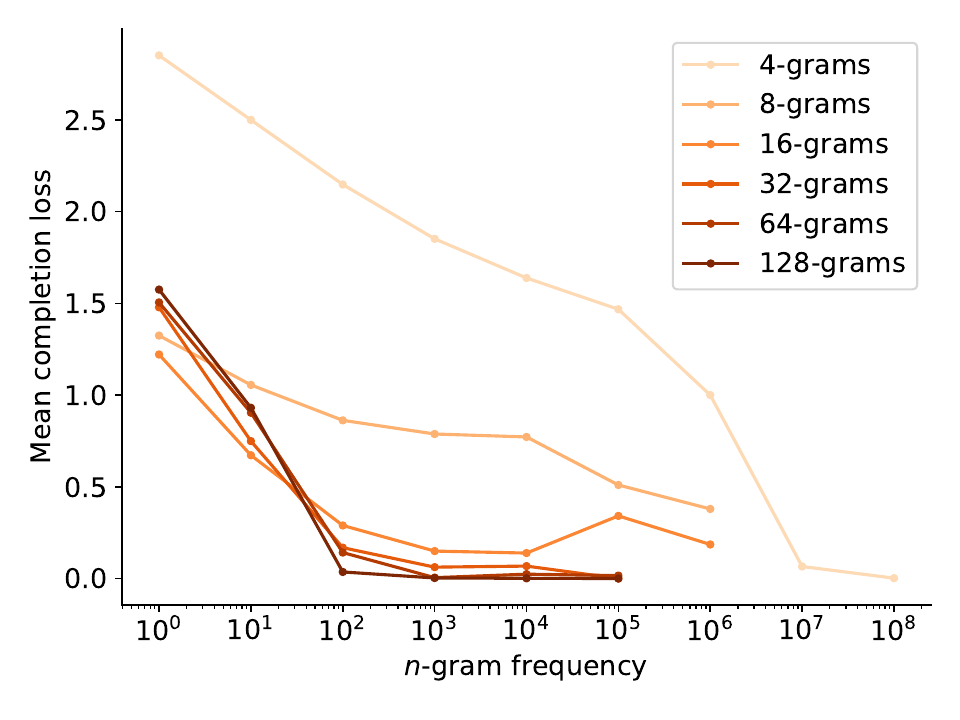}
        \caption{Completion loss as a function of $n$-gram training frequency for different $n$-gram sizes. Across $n$-gram sizes, more frequent $n$-grams have lower loss (with larger $n$ being easier to predict).}
        \label{fig:cl-frequency}
    \end{subfigure}
    \caption{$n$-gram completion loss based on presence in train and frequency.}
\end{figure*}

\paragraph{Training $n$-Grams are Easier to Complete.} \Cref{fig:cl-occurrence} shows that, across $n$-gram sizes, the completion loss for $n$-grams from the training set is smaller than for $n$-grams not in the training set (concretely, for $n$-grams above size 10, the completion loss is roughly 50\% less when the $n$-gram was in the training set). For $n > 80$, the loss curve for $n$-grams not in training becomes noisy, reflecting the rarity of such $n$-grams.
These results suggest that Pythia-12B is upweighting tokens that complete $n$-grams from pretraining.\footnote{While these results may be confounded (training $n$-grams may be easier to complete for other reasons besides their occurrence in the training set), we believe this is not a significant issue. See \Cref{sec:by-index} for analysis of a possible confounder: \textsc{nnsl} by token index.}
This finding potentially explains why more constrained decoding decreases novelty: while LMs assign probability to complete training $n$-grams, their next-token prediction with standard sampling also places a lot of probability mass on other tokens. Thus, training $n$-grams may not always get generated. However, the finding that training $n$-grams are upweighted in terms of probability suggests that pruning probability mass on other tokens (as approaches like top-$p$ or top-$k$ do) would cause even more training $n$-grams to be generated, as found in \Cref{sec:relative-novelty}.

\paragraph{Frequent $n$-Grams are Easier to Complete.} \Cref{fig:cl-frequency} shows that, across sizes, $n$-grams that are more frequent in the training data are easier for Pythia-12B to complete, implying LM predictions are sensitive to training data frequency effects.
This is particularly relevant when specific token continuations are compared to assess multiple choice answers: e.g., \texttt{a)}, \texttt{b)}, \texttt{c)}, and \texttt{d)} for MMLU evaluation \citep{hendryckstest2021}, or comparing \texttt{is}/\texttt{are} to assess noun-verb agreement competence \citep{marvin-linzen-2018-targeted}.
\Cref{tab:ngram-frequencies} shows that the Pile frequency of these continuations are not uniform. Combined with \Cref{fig:cl-frequency}, this suggests evaluating LMs by comparing these tokens may be susceptible to pretraining frequency effects.

In summary, these results suggest that LMs are more likely to memorize and copy $n$-grams with higher frequency in the pretraining data.

\begin{table}[t!]
    \centering
    \resizebox{0.8\columnwidth}{!}{
    \begin{tabular}{cccc}
        \toprule
         \texttt{a)} & \texttt{b)} & \texttt{c)} & \texttt{d)} \\
         \underline{$2.5 \times 10^7$} & $2.3 \times 10^7$ & $2.1 \times 10^7$ & $1.1 \times 10^7$ \\
         \midrule
         & \texttt{is} & \texttt{are} & \\
         & \underline{$1.4 \times 10^8$} & $2.9 \times 10^7$ & \\
         \bottomrule
    \end{tabular}
    }
    \caption{$n$-gram frequencies in the Pile computed by CDAWG. \texttt{a)} is more frequent than other options, and \texttt{is} is more frequent than \texttt{are}. Combined with \Cref{fig:cl-frequency}, this suggests evaluations that use continuation probabilities may be susceptible to pretraining frequency effects.}
    \label{tab:ngram-frequencies}
\end{table}

\section{Related Work}

\subsection{Methods for Accessing Text Corpora}

Data is becoming an important factor for understanding LM behavior \citep{elazar2023s}. As the scale of pretraining datasets continues to increase, naive search through these large datasets does not scale. As such, we need clever algorithms and data structures to interact with and study huge datasets.

\citet{mccoy2021language}, the first work to study the generation novelty of LMs, trained on Wikitext-103 (<1 GB) and WebText (40 GB). At this small data scale, they matched $n$-grams by brute-force enumeration, which would not be feasible today with the Pile (1254 GB) or larger datasets.
In contrast, \citet{elazar2023s} use an elastic search index based on an inverted index that allows a to search a corpus which depends on the number of documents in the corpus, making it much slower then our approach.
\citet{carlini2023quantifying,liu2024infinigram} use a suffix array \citep{suffix-arrays}, allowing queries in logarithmic time w.r.t.~corpus size.
Another data structure previously used in the setting of text generation with retrieval is the FM-index \citep{fm-index,bevilacqua2022autoregressive}, a compressed suffix array.

In this work, we use a CDAWG \citep{Crochemore1997,inenaga2005online}, which is a refinement of the earlier DAWG \citep{blumer1984building}, and part of a larger family of ``$*$DAWG'' indices \citep{takagi2018linear,inenaga2024linearsize}. $*$DAWGS use more memory than suffix arrays but support faster membership and \textsc{nnsl} queries (cf.~\Cref{sec:cdawgs}). $*$DAWGs also naturally allow computing arbitrary-length $n$-gram probabilities \citep{liu2024infinigram}, which could be useful for retrieval language modeling applications.

\subsection{Memorization, Contamination, and Generalization}

The increased use of LMs has raised concerns about memorization artifacts that might limit their generalization potential. For instance, \citet{bender2021dangers} draw a parallel of LMs to ``stochastic parrots'' that memorize and mimic their training data. 

Memorization has been carefully studied and quantified \citep{zhang2021understanding,kandpal2022deduplicating,lee2022deduplicating,magar2022data,carlini2023quantifying,ippolito-etal-2023-preventing} and is often framed as a concerning property of model behavior.
On the other hand, other works claim that memorization is integral for generalization \citep{feldman2020does,feldman2020neural,chatterjee2018learning}.
In this work, we do not take a stance on the importance or dangers of memorization, but rather quantify the novelty of LM-generated text vs. human text and investigate how different parameters affect novelty.
In contrast to much previous work on memorization, we also focus on the novelty of typical text rather than text elicited in adversarial settings \citep{carlini2023quantifying,ippolito-etal-2023-preventing}.


Like \citet{mccoy2021language}, we focus on generation novelty rather than quality, and we are interested in the effect of different variables such as model size, and decoding strategies on the generation novelty. Due to the CDAWG, we are able to scale our analysis to larger datasets than \citet{mccoy2021language}. In addition to text diversity, \citet{shaib2024standardizing,syntactic-templates} investigated the diversity of generated part-of-speech sequences rather than texts, as an abstract measurement over the raw texts.
\citet{lesci-etal-2024-causal} consider the effect of data ordering on LM memorization.
In future work, it would be possible to extend our CDAWG analysis to look at data ordering by building the CDAWG on documents in the same ordering as a pretraining run and performing analysis at different checkpoints of the partial CDAWG.




\section{Discussion}
\paragraph{A Note on Generation Quality.} 
In this work, we focus on quantifying $n$-gram novelty in model generations with respect to the model's training data. One confounding factor that may influence the \textit{cause} of novelty in such generations is the quality of the text. For example, consider a model that generates the sequence ``n-gram n-gram n-gram n-gram.'' This sequence is of low quality and is unlikely to appear in a corpus (it does not appear in Dolma), making it a novel 4-gram. Investigating how $n$-gram novelty interacts with text quality is left for future work.

\paragraph{Other CDAWG Use Cases.}
Several recent studies have explored memorization during training \cite{chang2024large,lesci-etal-2024-causal}. These studies performed interventions on the data used for training LMs at different checkpoints to study the effect of model size, training step, etc. on memorization.
Since CDAWG is constructed sequentially, future research could leverage this property to mirror the order in which the data was presented to the model during training. This would enable the use of CDAWGs for studying
how novelty evolves over the course of pretraining for a specific data ordering.

\section{Conclusion}

We introduce \textsc{Rusty-DAWG}, an efficient index for finding \emph{unbounded length} $n$-gram overlap against a pretraining corpus whose runtime is independent of the corpus size. Using \textsc{Rusty-DAWG}, we show that, at large $n$, Pythia generates less novel $n$-grams than novel human-written text.
We also find that increasing model size, constrained decoding (e.g., with temperature 0), or prompting with training data decrease novelty --- in particular, low temperature has a dramatic effect.
Finally, more frequent training $n$-grams are completed by LMs with \emph{lower loss}, suggesting LMs are more prone to memorizing frequent $n$-grams.
We hope \textsc{Rusty-DAWG} enables further analysis of pretraining data as well as decontamination \citep{paloma} and retrieval language modeling \citep{khandelwal2020generalization,liu2024infinigram} research.

\section*{Limitations}
When evaluating novelty, we focus on verbatim $n$-gram novelty rather than evaluating semantic novelty, which would be harder to operationalize.
Our analysis focuses on the non-deduplicated Pile, a primarily English dataset.
There are many variables about data curation or LM training that could affect generation novelty beyond the ones we have considered, which could be explored using similar methodology in future work.
Finally, as discussed in \Cref{sec:cdawg-memory}, one challenge with deploying the CDAWG is the memory overhead, though we believe this can be optimized in future work.

\section*{Acknowledgements}
We thank Ananya Jha, Rodney Kinney, Dave Wadden, and Pete Walsh for their engineering contributions to \textsc{Rusty-DAWG} during the AI2 hackathon and Michal Guerquin, Johann Dahm, and the Beaker team for assistance running \textsc{Rusty-DAWG} on AI2 infrastructure.
We thank Shunsuke Inenaga for guidance on the CDAWG data structure and Cyril Allauzen, Jiacheng Liu, and Vishakh Padmakumar for advice and feedback.
Thanks to Ian Magnusson for finding a typo in \Cref{fig:example-cdawg}.

\bibliography{references}

\clearpage
\appendix
\section{Computing $n$-Novelty from \textsc{nnsl}} \label{sec:computing-novelty}

The direct output of the CDAWG is $L_Q$, the \textsc{nnsl} vector across each position in $Q$.
We now describe how to compute the $n$-novelty curve from $L(Q)$.
First, we define $c(n)$ as the the number of times $n$ occurred in $L(Q)$:
\begin{equation*}
    c(n) = \sum_{\ell \in L(Q)} \mathbbm{1}[n = \ell] .
\end{equation*}
Next, the number of novel $n$-grams in $Q$ is
\begin{equation*}
    \left(\sum_{n' < n} c(n') \right) - (n - 1) .
\end{equation*}
The total number of $n$-grams in $Q$ is $\abs{Q} - (n - 1)$.
Thus, the $n$-novelty is
\begin{equation*}
    n\mathrm{-novelty}(Q) = \frac{\left(\sum_{n' < n} c(n') \right) - (n - 1)}{\abs{Q} - (n - 1)} .
\end{equation*}
This can be extended to multiple documents by summing the numerator and denominator across documents before dividing.

\section{CDAWG Details} \label{app:cdawgs}

\subsection{Querying the CDAWG} \label{sec:cdawg-query}

\Cref{alg:query-cdawg} implements an \textsc{nnsl} query by greedily passing $Q$ through the CDAWG one token at a time. We track the current state, any intermediate progress along an arc represented by indices $\langle \alpha, \gamma \rangle$ for a span in $C$, and the currently matched length. If no progress can be made along an arc by the next token, a failure arc \citep{pmlr-v217-allauzen23a} is followed to back off until a state with a defined transition is found (or to $\emptyset$ if no such state exists).
If some partial progress is matched along an arc, that progress must be matched at the arc out of the state backed off to as well.
We refer to this as an \emph{implicit failure transition}, denoted by $\phi(q, \langle \alpha, \omega \rangle)$.

\begin{algorithm}[t]
\SetNoFillComment
\caption{\textsc{nnsl} query with CDAWG}\label{alg:query-cdawg}
\SetKwInOut{Input}{Input}
\SetKwInOut{Output}{Output}
\KwData{CDAWG $G$ with source $q_0$}
\KwIn{query $Q$}
\KwOut{\textsc{nnsl} vector $L_Q$ and counts $N_Q$, emitted pairwise}

$s.q \gets q_0$\;
$s.\langle\alpha, \omega\rangle \gets \langle0, 0\rangle$\;
$s.\ell \gets 0$\;
\For{token $\sigma$ of $Q$}{
    $s \gets \texttt{trans}(s, \sigma)$\;
    \textbf{emit} $\langle s.\ell, \mathsf{count}(s.q)\rangle$\;
}

\SetKwFunction{FMain}{trans}
\SetKwProg{Pn}{fn}{:}{\KwRet}
\Pn{\FMain{$s$, $\sigma$}}{
    \uIf{$s.q = \emptyset$}{
        $s.q \gets q_0$\;
        $s.\langle\alpha, \omega\rangle \gets \langle0, 0\rangle$\;
        $s.\ell \gets 0$\;
    }
    \uElseIf{$\alpha = \omega$}{
        $q' \gets$ target of completed arc\;
        \eIf{$e \gets \sigma$-edge out of $q'$} {
            $s.q \gets q'$\;
            $s.\langle\alpha, \omega\rangle \gets$ weight of $e$\;
            $s.\ell \gets s.\ell + 1$\;
        }{
            $s.q \gets \phi(q', \langle \alpha, \omega \rangle)$\;
            $s \gets \texttt{trans}(s, \sigma)$\;
        }
    }
    \uElse{
        $\sigma' \gets$ token $s.\alpha$ of $C$\;
        \eIf{$\sigma = \sigma'$}{
            $s.\alpha \gets s.\alpha + 1$\;
            $s.\ell \gets s.\ell + 1$\;
        }{
            $s.q \gets \phi(q, \langle \alpha, \omega \rangle, \ell)$\;
            $s \gets \texttt{trans}(s, \sigma)$\;
        }
    }
    \Return $s$\;
}
\end{algorithm}

\subsection{Populating Counts} \label{sec:cdawg-counts}

We build the CDAWG according to Figure 17 of \citet{inenaga2005online}. The final post-processing step we add is to populate the counts in the CDAWG via a depth-first traversal (cf.~\Cref{alg:add-counts}).
The idea is that the CDAWG represents the frequency of a string $x$ in $C$ by the number of paths from the node reached by $x$ to a sink node. Further, the frequency of each node is the sum of the frequencies of its children. Thus, we can populate all the counts in the CDAWG via a depth-first traversal of its nodes, which takes time $O(\abs{C})$. 

\begin{algorithm}[t]
\SetNoFillComment
\caption{Add counts to CDAWG}\label{alg:add-counts}
\KwData{CDAWG $G$ with source $q_0$}
create stack $S$\;
push $\langle\open, q_0\rangle$ onto $S$\;
\While{$\langle o, q\rangle \gets$ pop from $S$}{
    \eIf{$o = \open$}{
        \If{$\mathsf{count}(q) > 0$}{
            \Continue;
        }
        $\mathsf{count}(q) \gets 1$\;
        push $\langle\close, q\rangle$ onto $S$\;
        \For{child $q'$ of $q$}{
            push $\langle\open, q'\rangle$ onto $S$\;
        }
    }{
        $\mathsf{count}(q) \gets 0$\;
        \For{child $q'$ of $q$}{
            add $\mathsf{count}(q')$ to $\mathsf{count}(q)$\;
        }
    }
}
\end{algorithm}

\subsection{Graph Representation} \label{sec:cdawg-memory}

An important detail for the memory usage of the CDAWG is it is represented as a graph. We represent the graph as a list of nodes and a list of edges. The edges at each node are represented by a binary AVL tree, which means the arc labelled by token $\sigma \in \Sigma$ can be found in $O(\log \abs{\Sigma})$ time. However, this representation means each edge takes 26 bytes (with 5 byte pointers), which leads to an overall size of $29 \abs{C}$ for the CDAWG. This is roughly 4${\times}$ larger than the corresponding suffix array, meaning there is a time/space tradeoff between the two approaches. We believe the memory overhead factor of the CDAWG could be significantly optimized by refining this graph representation in future work.

The memory overhead of \textsc{Rusty-DAWG} could be further reduced by implementing recent improvements of the CDAWG such as the linear-size CDAWG (LCDAWG; \citealp{takagi2018linear}) and simplified LCDAWG (simLCDAWG; \citealp{inenaga2024linearsize}).

\subsection{\textsc{Rusty-DAWG} Library} \label{sec:cdawg-library}

DAWGs and CDAWGs can be stored in either RAM or disk to accomodate different resource constraints (building and inference are faster in RAM, but for very large datasets, using disk may be preferable due to resource constraints, especially for inference).
Rust was chosen as a language so runtime and memory overhead could be optimized, though we also created Python bindings for easy integration with machine learning workflows.
All experiments in the paper were carried out using the Python bindings to access a CDAWG built with the \textsc{Rusty-DAWG} library.

While building our CDAWGs for the Pile, we stored them in RAM to make the process faster.
This took 24 hours using 30 Google Cloud \texttt{n2-highmem-48} machines.
At inference time, we transferred the CDAWGs to disk.
The \textsc{Rusty-DAWG} inference speed depends mainly on the speed of the RAM or disk used to store the data structure since inference is bottlenecked by memory accesses. We used relatively cheap \texttt{n2-standard-16} machines with low IO speed (15k IOPS) and observed between 100-1000 tokens per second, which sufficed for our experiments. Since memory accesses the bottleneck, the speed could be improved significantly just by using machines with faster disk (or RAM).

\section{Lower Bound on Novelty Without Duplication} \label{sec:theoretical-lb}

Our theoretical lower bound baseline is based on the idea that the next token is fundamentally nondeterministic, and, therefore, long $n$-gram spans should be unlikely.

\subsection{Warmup: Always Nondeterministic Case}

Let $\mathbbm{1}[\cdot]$ be an indicator, and we will use subscripts (e.g., $X_i$ or $d_i$) to indicate elements of strings.

Say that we sample a corpus $C$ of strings from some distribution $p$ and then denote by $\mathcal D_n$ the set of all $n$-grams in $C$.
We then let $X$ be a random string of length $n$ sampled from  $p$. 
We say that $X$ is $n$-novel if $X \not\in \mathcal D_n$ and we are interested in analyzing this probability. 
This event is the complement of the events where $X$ is any particular $n$-gram from $C$, so its probability is
\begin{equation*}
    p(X \; n\mathrm{-novel}) = 1 - p\left(\bigvee_{d \in \mathcal D_n} \bigwedge_{i=1}^n \mathbbm{1}[ X_i = d_i ]\right) .
\end{equation*}
By the union bound,
\begin{align*}
    p(X \; n\mathrm{-novel})
    &\geq 1 - \sum_{d \in \mathcal D_n} p\left( \bigwedge_{i=1}^n \mathbbm{1}[ X_i = d_i ]\right) \\
    &= 1 - \sum_{d \in \mathcal D_n} \prod_{i=1}^n p(X_i = d_i \mid X_{<i}) .
\end{align*}

\begin{figure*}[t]
    \centering
    \includegraphics[width=.3\textwidth]{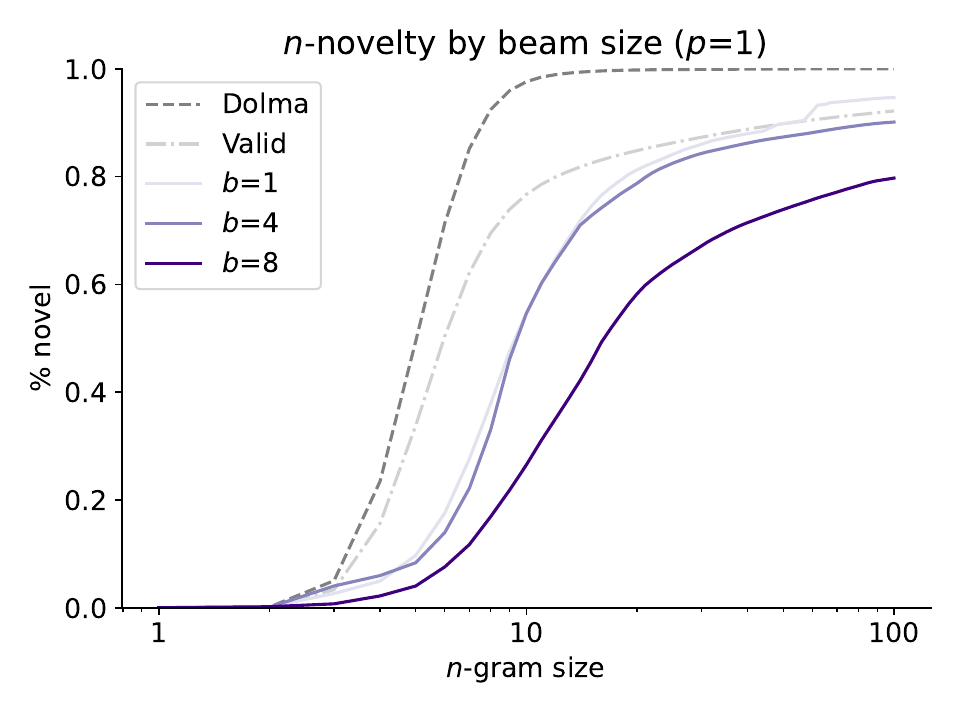}
    \includegraphics[width=.3\textwidth]{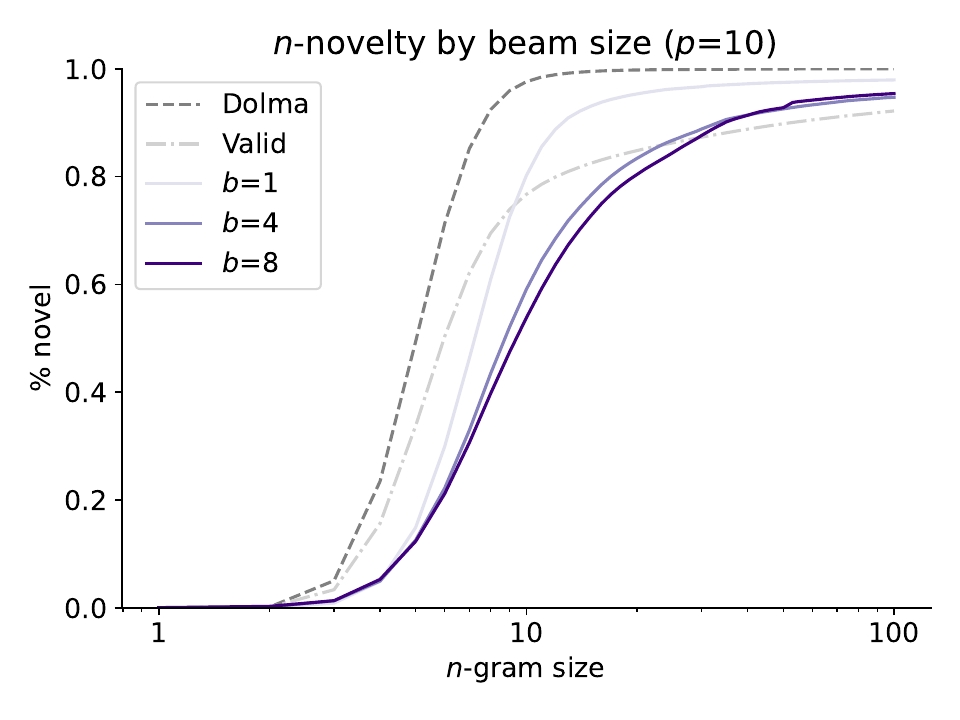}
    \includegraphics[width=.3\textwidth]{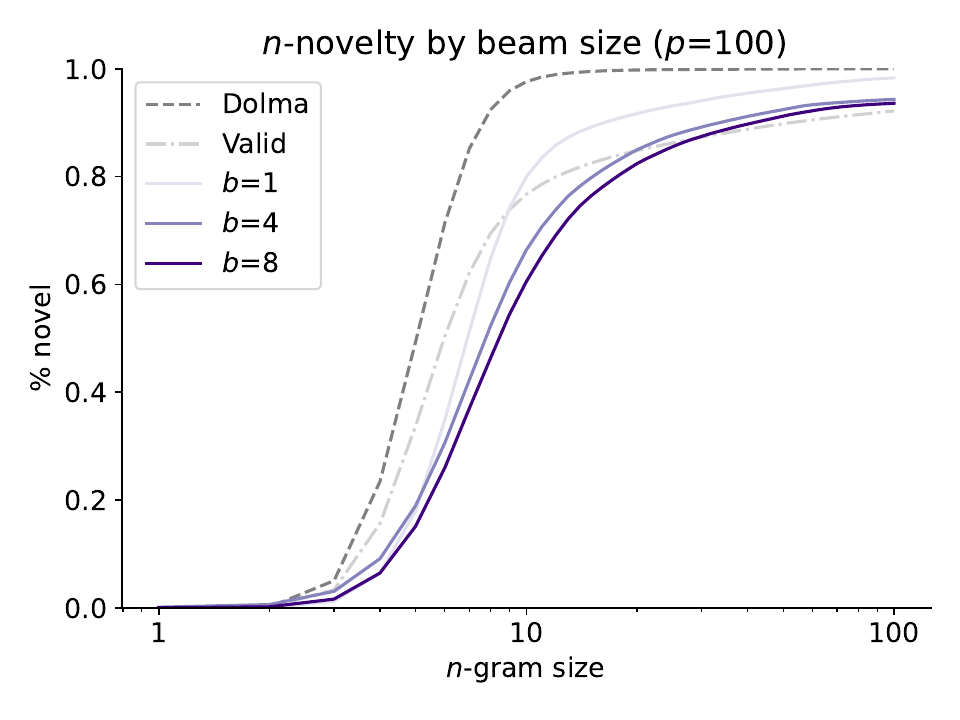}

    \caption{Beam decoding results with different amounts of training tokens used as a prompt: 1 token (left), 10 tokens (center), and 100 tokens (right).}
    \label{fig:beam-search-plen}
\end{figure*}

Assume $p$ is always nondeterministic at every position, so there is some $q < 1$ such that, for all $i$ and $x$,
\begin{equation} \label{eq:q-assumption}
    p(X_i = x_i \mid X_{<i} = x_{<i}) \leq q .
\end{equation}
\noindent Then it follows that
\begin{align*}
    p(X \; n\mathrm{-novel})
    &\geq 1 - \sum_{d \in \mathcal D_n} \cdot q^n \tag{by \ref{eq:q-assumption}} \\
    &= 1 - \abs{\mathcal D_n} \cdot q^n \\
    &\geq 1 - \abs{C} \cdot q^n .
\end{align*}
Finally, recasting in terms of negative log-likelihood $\ell = - \log q > 0$, we get
\begin{equation*}
    p(X \; n\mathrm{-novel}) \geq 1 - \abs{C} \cdot \exp(-n\ell) .
\end{equation*}
A first observation here is that $p(X \; n\mathrm{-novel})$ should exponentially decay $1$ quickly with $n$.


\begin{figure*}
    \centering
    \begin{subfigure}[t]{0.45\linewidth}
        \includegraphics[width=\columnwidth]{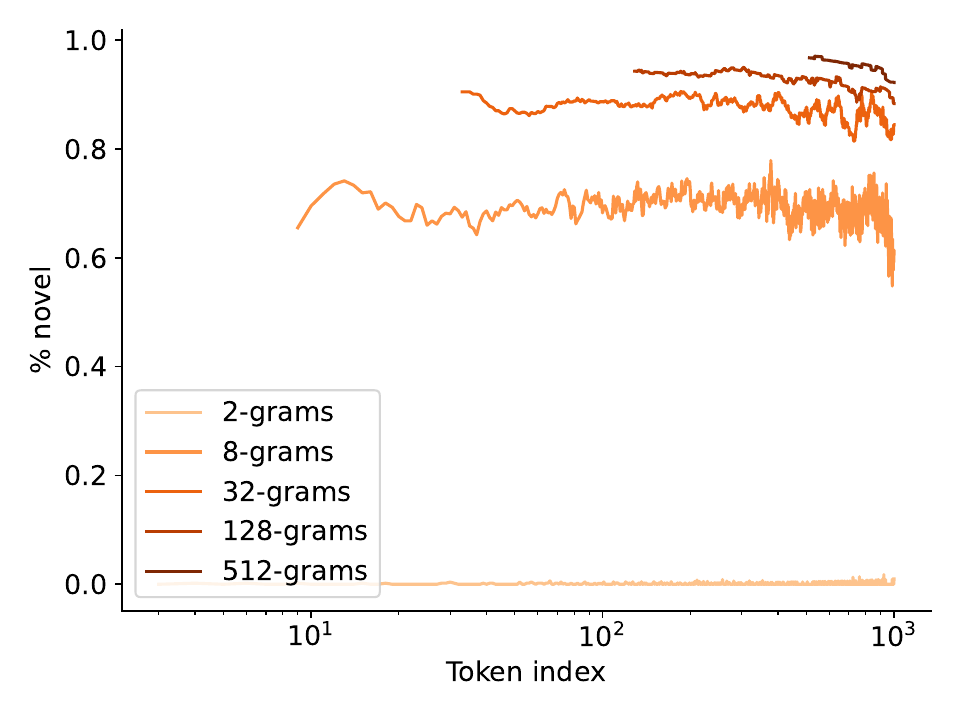}
        \caption{Validation}
        \label{fig:by-index-val}
    \end{subfigure}
    \begin{subfigure}[t]{0.45\linewidth}
        \includegraphics[width=\columnwidth]{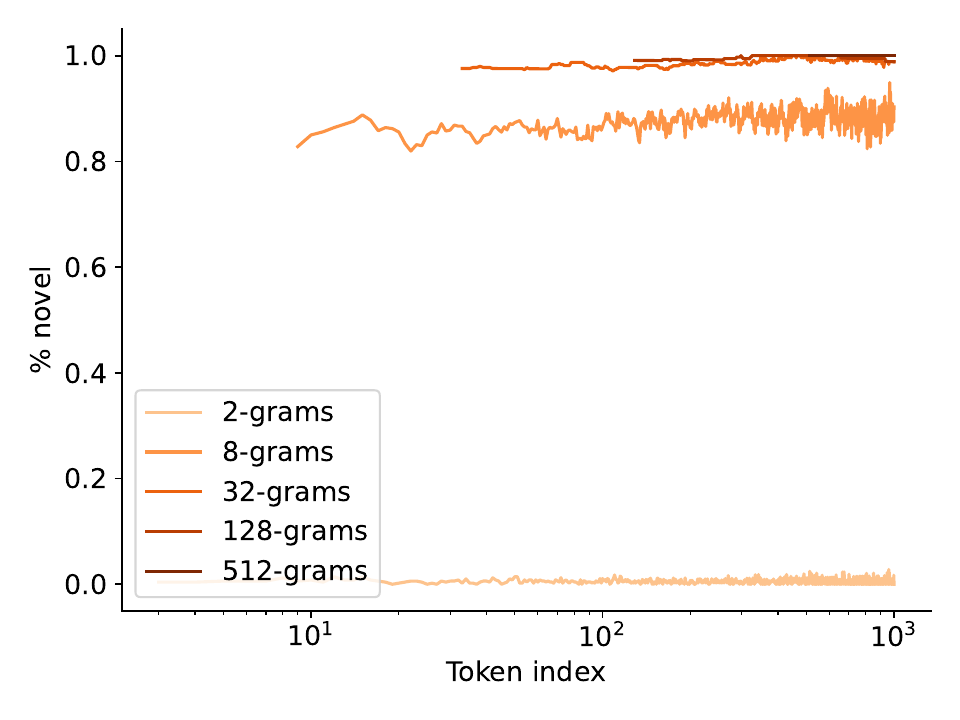}
        \caption{Pythia-12B}
        \label{fig:by-index-pythia}
    \end{subfigure}
    \caption{$n$-novelty by token index in validation and Pythia-generated text.}
    \label{fig:by-index}
\end{figure*}

\subsection{Probably Nondeterministic Case}

The previous derivation assumes that all tokens are nondeterministic.
We relax this slightly by assuming that, within an $n$-gram, some tokens can be deterministic, but some fixed rate $r$ will not be.
More formally, we assume there exists $r$ such that, for all $x$, $p(X_i = x_i \mid X_{<i} = x_{<i}) \leq q$ with probability $r$ (over $i$).
This implies that the novelty is as follows:

\begin{align*}
    p(X \; n\mathrm{-novel})
    &\geq  1 - \sum_{d \in \mathcal D_n} \prod_{i=1}^n p(X_i = d_i \mid X_{<i}) \\
    &\geq 1 - \sum_{d \in \mathcal D_n} q^{rn} \cdot 1^{(1-r)n}\\
    &= 1 - \sum_{d \in \mathcal D_n} q^{rn} \\
    &= 1 - \abs{C} \exp(n (\log r - \ell)) .
\end{align*}



This is the form of the lower bound invoked in the main plots (cf.~\Cref{sec:baselines}).
We believe this assumption is possibly strong, but it allows us to get a first-pass lower bound.

\section{Beam Search Results Elaboration}
\label{app:beam-novelty}

The beam search decoding used in \Cref{fig:by-decoding} is deterministic because the temperature is 0 and the prompt is null. To complement these results, we also include additional results in \Cref{fig:beam-search-plen} where a prompt of length $p$ (taken from the training data) is used. In this regime, we find that, similar to the promptless results, beam search decreases $n$-novelty. However, the novelty curve is not so extreme for beam size 8.
This indicates that, with beam size 8, the LM does not always copy very large chunks of training documents like in \Cref{fig:by-decoding}.

\wm{A cleaner experiment here could be to prompt with validation or Dolma data rather than training data, or just to increase the temperature, if we just want to induce nondeterminism.}

\section{Novelty by Token Index} \label{sec:by-index}

We also present an additional analysis showing how novelty varies across token index in the document, for both validation and LM-generated text. As shown in \Cref{fig:by-index}, $n$-novelty stays roughly constant across different token positions across both validation and LM-generated text, for every value of $n$. This pattern in the validation text in particular (\Cref{fig:by-index-val}) suggests token index does not correlate with $n$-gram size, and is thus likely not a confounder in \Cref{sec:completion-loss}.

\end{document}